\documentclass[sigconf]{acmart}

\usepackage{url}
\usepackage[utf8]{inputenc} 
\usepackage[T1]{fontenc}

\usepackage{hyperref}       
\usepackage{url}            
\usepackage{booktabs}       
\usepackage{amsfonts}       
\usepackage{nicefrac}       
\usepackage{microtype}      
\usepackage{xcolor}         
\usepackage{amsmath}
\usepackage{mathtools}
\usepackage{amsthm}
\usepackage{multirow}
\usepackage{makecell}
\usepackage[ruled,vlined,linesnumbered]{algorithm2e}
\usepackage{microtype}
\usepackage{graphicx}
\usepackage{subfigure}
\usepackage{booktabs} 
\usepackage{array}
\usepackage{textcomp}
\usepackage{stfloats}
\usepackage{verbatim}
\usepackage{balance}
\usepackage{caption}
\usepackage{tabularx}
\usepackage{siunitx}
\AtBeginDocument{%
  }

\setcopyright{acmlicensed}
\copyrightyear{2025}
\acmYear{2025}
\acmDOI{XXXXXXX.XXXXXXX}
\acmConference[Conference acronym 'XX]{Make sure to enter the correct
  conference title from your rights confirmation email}{
  2025}{Woodstock, NY}
\acmISBN{978-1-4503-XXXX-X/2018/06}

\begin{document}

\title{TriForecaster: A Mixture of Experts Framework for Multi-Region Electric Load Forecasting with Tri-dimensional Specialization}

\author{Zhaoyang Zhu}
\email{zhuzhaoyang.zzy@alibaba-inc.com}
\authornote{Contributed equally to this research.}
\orcid{0000-000x-xxxx-xxxx}
\affiliation{%
  \institution{DAMO Academy, Alibaba Group}
  \streetaddress{969 West Wen Yi Road, Yu Hang District}
  \city{Hangzhou}
  \state{Zhejiang}
  \country{China}
  \postcode{311121}
}

\author{Zhipeng Zeng$^*$}
\email{zengzhipeng.zzp@alibaba-inc.com}
\orcid{0000-000x-xxxx-xxxx}
\affiliation{%
  \institution{DAMO Academy, Alibaba Group}
  \streetaddress{969 West Wen Yi Road, Yu Hang District}
  \city{Hangzhou}
  \state{Zhejiang}
  \country{China}
  \postcode{311121}
}

\author{Qiming Chen$^*$}
\email{chenqiming.cqm@alibaba-inc.com}
\orcid{0000-000x-xxxx-xxxx}
\affiliation{%
  \institution{DAMO Academy, Alibaba Group}
  \streetaddress{969 West Wen Yi Road, Yu Hang District}
  \city{Hangzhou}
  \state{Zhejiang}
  \country{China}
  \postcode{311121}
}

\author{Linxiao Yang}
\email{linxiao.ylx@alibaba-inc.com}
\orcid{0000-0001-9558-7163}
\affiliation{%
  \institution{DAMO Academy, Alibaba Group}
  \streetaddress{969 West Wen Yi Road, Yu Hang District}
  \city{Hangzhou}
  \state{Zhejiang}
  \country{China}
  \postcode{311121}
}

\author{Peiyuan Liu}
\email{liupeiyuan.lpy@alibaba-inc.com}
\orcid{0000-0001-9558-7163}
\affiliation{%
  \institution{DAMO Academy, Alibaba Group}
  \streetaddress{969 West Wen Yi Road, Yu Hang District}
  \city{Hangzhou}
  \state{Zhejiang}
  \country{China}
  \postcode{311121}
}

\author{Weiqi Chen}
\email{jarvus.cwq@alibaba-inc.com}
\orcid{0000-0001-9558-7163}
\affiliation{%
  \institution{DAMO Academy, Alibaba Group}
  \streetaddress{969 West Wen Yi Road, Yu Hang District}
  \city{Hangzhou}
  \state{Zhejiang}
  \country{China}
  \postcode{311121}
}

\author{Liang Sun}
\email{liang.sun@alibaba-inc.com}
\orcid{0009-0002-5835-7259}
\affiliation{%
  \institution{DAMO Academy, Alibaba Group}
  \streetaddress{969 West Wen Yi Road, Yu Hang District}
  \city{Hangzhou}
  \state{Zhejiang}
  \country{China}
  \postcode{311121}
}


\renewcommand{\shortauthors}{Zhu et al.}

\begin{abstract}
Electric load forecasting is pivotal for power system operation, planning and decision-making. 
The rise of smart grids and meters has provided more detailed and high-quality load data at multiple levels of granularity, from home to bus and cities. 
Motivated by similar patterns of loads across different cities in a province in eastern China, in this paper we focus on the Multi-Region Electric Load Forecasting (MRELF) problem, targeting accurate short-term load forecasting for multiple sub-regions within a large region. 
We identify three challenges for MRELF, including regional variation, contextual variation, and temporal variation. 
To address them, we propose TriForecaster, a new framework leveraging the Mixture of Experts (MoE) approach within a Multi-Task Learning (MTL) paradigm to overcome these challenges. TriForecaster features RegionMixer and Context-Time Specializer (CTSpecializer) layers, enabling dynamic cooperation and specialization of expert models across regional, contextual, and temporal dimensions. Based on evaluation on four real-world MRELF datasets with varied granularity, TriForecaster outperforms state-of-the-art models by achieving an average forecast error reduction of 22.4\%, thereby demonstrating its flexibility and broad applicability. 
In particular, the deployment of TriForecaster on the eForecaster platform in eastern China exemplifies its practical utility, effectively providing city-level, short-term load forecasts for 17 cities, supporting a population exceeding 110 million and daily electricity usage over 100 gigawatt-hours.
    
\end{abstract}





\maketitle

\section{Introduction}\label{sec:intro}

Electric load forecasting~\cite{ELF:2017:survey} predicts the future electricity demand based on historical data and other external factors. 
It is crucial to balance supply and demand of power, reduce costs, and integrate renewable energy.
Although the electric load forecasting shares some similarities with multivariate time series forecasting, it is generally more challenging. 
Key difficulties include the wide range of factors that influence temperatures, humidity, levels of economic activity and holiday effects, which are often only partially observable, making it difficult to capture all relevant variables in a comprehensive way. 
Furthermore, the impact of these factors exhibits considerable regional and temporal heterogeneity, resulting in distributional shifts that complicate the precision of the prediction. The forecasting process is further impeded by unexpected uncertainty factors, such as abrupt policy changes (e.g., environmental production restrictions), data quality issues arising from sensor malfunctions and transmission problems in smart meters, and the diversification of load structures driven by dual-carbon objectives. 
These issues lead to more complex load behaviors and multidimensional couplings across environmental, social, and economic dimensions, resulting in a limited and fragmented training dataset characterized by noise and missing values.

Recently, the rise of smart grids, smart meters, and multi-energy systems has provided access to more detailed and high-quality load data. In this paper, we study the Multi-Region Electric Load Forecasting (MRELF)  problem~\cite{Wu:MRELF:2022,Fan:MREFL:2010}.  MRELF generates accurate short-term load predictions for multiple sub-regions within a large region.
Unlike the traditional single-region electric load forecasting at the aggregated level, it considers the similarity and differences of different regions to make the load prediction for multiple regions. 

\begin{figure}[!t]
    \centering
    \includegraphics[width=\linewidth]{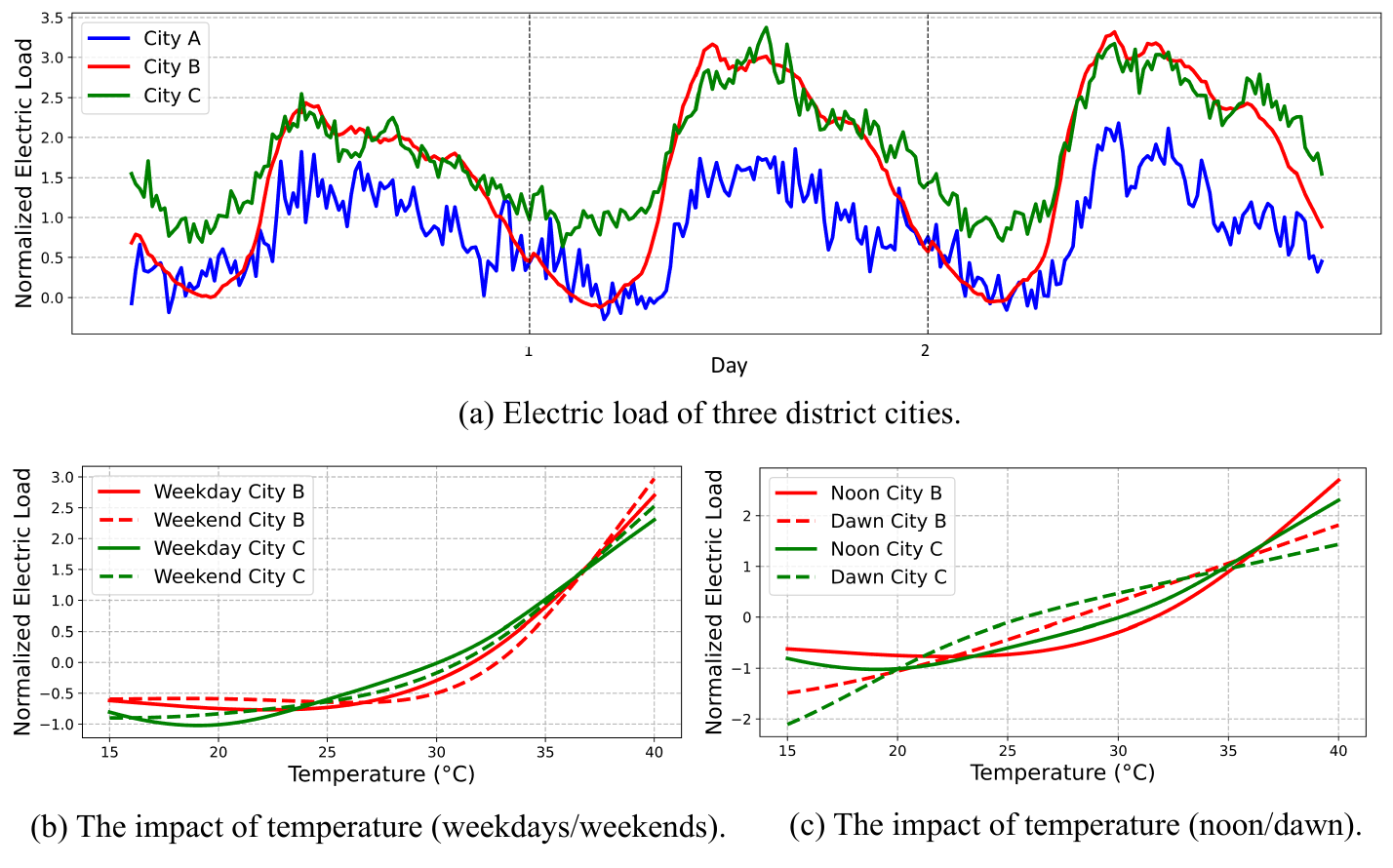}
    \caption{The motivation for addressing regional, contextual, temporal variations in multi-region electric load forecasting: (a) Different cities exhibit significant similarities in usage patterns, yet they differ markedly in scale and degree of fluctuation, namely Regional Variation; (b) Contextual Variation: Temperature impacts differ significantly between weekends and weekdays across different cities; (c) Temporal Variation: Temperature effects vary significantly between noon and dawn across different cities.}
    \label{fig:motivation_plot}
\end{figure}

Next we use real-world examples in a province (consisting of multiple cities) in eastern China to demonstrate typical patterns in MRELE, as illustrate in Figure~\ref{fig:motivation_plot}. 
A significant finding is the pronounced similarity in electric load patterns across these regions (cities in Figure~\ref{fig:motivation_plot}). This similarity arises from the fact that the electric load is a direct reflection of human activities, influenced by universal factors such as electricity pricing, daily routines, and weather conditions. 
Nevertheless, despite the similarities, the extent to which these factors impact the electric load varies significantly from three perspectives:
\begin{itemize}
    \item \textbf{Regional Variation}. This variation is largely attributable to regional differences in industrial structure and load composition. For instance, industrial hubs exhibit an increased sensitivity to electricity pricing, often adjusting production schedules in response to cost variations. In contrast, residential areas are more heavily influenced by weather.
    \item \textbf{Contextual Variation}. The impact of influencing factors on electric load is highly dependent on the specific values of other covariates, referred to as the context. Certain covariates may exert more influence in particular situations or contexts. For instance, the impact of covariates such as temperature on electric load during a hot, rainy weekday in summer can differ significantly from their impact on a mild, windy holiday. In this example, temperature plays the most important role, but other weather factors, season of the day, holiday/weekday represent the context. 
    \item \textbf{Temporal Variation}.  The influence of factors on electric load varies significantly over time throughout the day. For example, covariates such as temperature may have a lesser impact at dawn compared to their impact at noon on the same day.
\end{itemize}

To make accurate prediction for MRELE, multi-task learning (MTL)~\cite{fiot2016electricity, jiang2021hybrid, jiang2021hybrid, liu2023boosted} is wildly used in the literature.  MTL provides various techniques to facilitate the learning of both region-generic and region-specific components including methods based on Graph Neural Networks ~\citep{mansoor2024graph, chen2023research}.
However, ignoring the aforementioned regional, contextual, and temporal variations can lead to the gradient conflict problem~\citep{chen2020just, liu2021conflict} in MTL models, as samples from different areas and contexts may require distinct emphasis on various covariates, resulting in suboptimal performance and slower convergence rates. 
Recently, the development of Mixture of Experts (MoE) methods~\citep{chen2023mod, ye2023taskexpert, yang2024multi} within the MTL framework has shown significant potential in addressing gradient conflict issues among tasks. Despite their promises, the applicability of MoE methods to MRELF remains largely unexplored, as existing research primarily focuses on applications in the vision or language domain with abundant data and transformer-based architectures.

In this paper, we focus on the MRELF problem from the MTL perspective. To effective address the regional, contextual, and temporal variations aforementioned, we propose a new model, TriForecaster, which innovatively addresses the MRELF problem by using MoE within the MTL framework. TriForecaster enables flexible cooperation and specialization of experts across regional, contextual and temporal dimensions which further enhance the model's ability to leverage region-general information while simultaneously adapting to distinct regional, contextual, and temporal characteristics. Firstly, the model extracts and integrates both region-specific and region-generic information using stacks of RegionMixer layers. Within each layer, a shared general model captures region-generic features that are relevant across various regions, while distinct region-specific expert models focus on unique, localized attributes. The integration of these diverse features is achieved using a parameter-efficient stochastic fusion module, which intelligently combines information by assessing similarities between expert models ~\citep{han2024softs}. This method eliminates the need for region-specific gating networks and load balance auxiliary loss within each layer while leveraging both region-specific and aggregated information from all experts to enhance local adaptations\citep{han2024softs}. Subsequently, the integrated information within each layer undergoes processing through a series of Context-Time Specializer (CTSpecializer) layers to accommodate contextual and temporal variations. 
This sequential processing ensures comprehensive accommodation of contextual and temporal variations. Experimentally, TriForecaster demonstrates superior performance compared to state-of-the-art models on four real-world MRELF datasets collected from diverse global locations, with varying levels of granularity from city to bus level. These results underscore TriForecaster's exceptional performance, flexibility, and general applicability across different contexts and scales. 


We outline our contributions as follows.
\begin{enumerate}
    \item We introduce TriForecaster, a framework featuring the RegionMixer layer and the CTSpecializer layer, designed to enable flexible cooperation and specialization of experts across regional, contextual, and temporal dimensions, facilitated by a parameter-efficient stochastic fusion mechanism.
    \item The experimental results of four real world MRELF datasets from different locations around the world (including load forecasting from buses (small-scale, low-level) to cities (large-scale, high-level)) prove the most advanced performance, flexibility and versatility of TriForecast.
    \item TriForecaster has been integrated into our eForecaster platform~\citep{eForecaster2023} and deployed in a province in eastern China. It provides day-ahead short-term load forecasting for 17 cities, covering a population of over 110 million people and daily electricity usage exceeding 100 gigawatt-hours.
\end{enumerate}

\section{Related Work}\label{sec:background}
In this section, we briefly review methodologies in electric load forecasting and multi-region electric load forecasting.

\subsection{Electric Load Forecasting}
In traditional electric load forecasting, classic methods such as tree-based models~\citep{ke2017lightgbm, chen2016xgboost}, ARIMA~\citep{shumway2017arima}, and Prophet~\citep{taylor2018forecasting} have been widely utilized. While these methodologies are interpretable and offer insights, they often suffer from limited capacity in capturing the complex dependencies between electric load and various covariates~\citep{fida2024comprehensive}. To address these limitations, recent research has harnessed more powerful deep neural networks. Studies~\citep{oreshkin2021n, l2022transformer, abumohsen2023electrical, imani2021electrical, mounir2023short} have explored the application of Recurrent Neural Network (RNN)-based models~\citep{hochreiter1997long, chung2014empirical}, Convolutional Neural Network (CNN)-based models~\citep{bai2018empirical}, and Transformer-based models~\citep{zhou2022fedformer, chen2022learning, zhou2021informer}. Additionally, hybrid approaches~\citep{mounir2023short, cai2022short, asiri2024short} have been proposed to further enhance model capacity. Despite these advances, a significant challenge still remains: the efficacy of these models is often contingent upon the availability of large, high-quality datasets. This is particularly problematic in the domain of electric load forecasting, where the data for a single region is limited and the datasets are frequently characterized by noise and missing values, constraining the application of these sophisticated techniques.


\subsection{Multi-region Electric Load Forecasting}
A  na\"{\i}ve approach to MRELF involves training models independently for each region. However, this method overlooks the potential to leverage commonalities between regions and fails to make use of the enriched datasets that could enhance predictive accuracy. Alternatively, Multi-Task Learning (MTL) treats each region as a distinct task, allowing strategies~\cite{fiot2016electricity, jiang2021hybrid, jiang2021hybrid, liu2023boosted} that blend region-generic and region-specific factors. To further utilize shared knowledge, another category of methods~\citep{mansoor2024graph, chen2023research} leverages Graph Neural Networks (GNNs) to capture dependencies across different regions. However, these methods often demand manual customization, increase computational costs, and may suffer from gradient conflicts.

Emerging Mixture of Experts (MoE) methods within the MTL framework ~\citep{chen2023mod, ye2023taskexpert, yang2024multi} seek to address these issues by using multiple expert models and distinct gating networks for task-specific dynamic cooperation, but they come with significant computational overheads and potential specialization issues due to load-balancing constraints and task-specific gating networks.

In contrast to previous approaches, our proposed method seeks to establish a generally applicable model for MRELF. We utilize MoE to achieve specialization across regional, contextual, and temporal dimensions, eliminating the need for task-specific gating networks and artificial load-balance auxiliary losses.

\section{Problem Formulation}

  We first briefly introduce electric load forecasting and multi-region electric load forecasting.


\textbf{Electic Load forecasting.}\hspace{0.5mm} Electricity load forecasting extends beyond historical data, incorporating future covariates such as weather forecasts and date features that influence electricity consumption. Given historical observations $\mathbf{S} \in \mathbb{R}^{L \times C}$ and future covariates $\mathbf{Z} \in \mathbb{R}^{H \times C_z}$, denoted collectively as $\mathbf{X} = (\mathbf{S}, \mathbf{Z})$, the objective is to learn a predictive model $F_{\boldsymbol{\theta}}$, which aims to forecast future electricity load demands $\mathbf{y} \in \mathbb{R}^{H \times 1}$, formalized as: $\mathbf{X} \xrightarrow[]{{F}(\cdot;{\boldsymbol{\theta}})}  \mathbf{y}$.

\textbf{Multi-region Electric Load Forecasting via MTL.}\hspace{0.5mm} 
Assume there are $T$ different regions, MRELF seeks to simultaneously predict future electric load for these regions. This is conceptualized as a MTL problem, where each region is defined as a distinct task $\{\mathcal{T}_1, \ldots, \mathcal{T}_T\}$. For each region $t$, there is an associated dataset $\mathcal{D}_t = \{(\mathbf{X}^t_i, \mathbf{y}^t_i)\}_{i=1}^{N_t}$, where $\mathbf{X}^t_i$ represents the composite input features, including historical observations and future covariates, and $\mathbf{y}^t_i$ denotes the target load demand. The objective is to learn a predictive model $F(\cdot)$ using a shared set of parameters $\boldsymbol{\theta}_s$ applicable across all regions, as well as region-specific parameters $\{\boldsymbol{\theta}_t\}_{t=1}^{T}$ for each individual task. The learning problem can be formulated as an optimization problem, aiming to minimize the cumulative loss across all regions:

\[
\min_{\boldsymbol{\theta}_s, \{\boldsymbol{\theta}_t\}_{t=1}^{T}} \sum_{t=1}^{T} \frac{1}{N_t}\sum_{i=1}^{N_t} \mathcal{L}(F(\mathbf{X}^t_i; \boldsymbol{\theta}_s, \boldsymbol{\theta}_t), \mathbf{y}^t_i).
\]

\section{TriForecaster}

\begin{figure*}[!t]
    \centering
    \includegraphics[width=\linewidth]{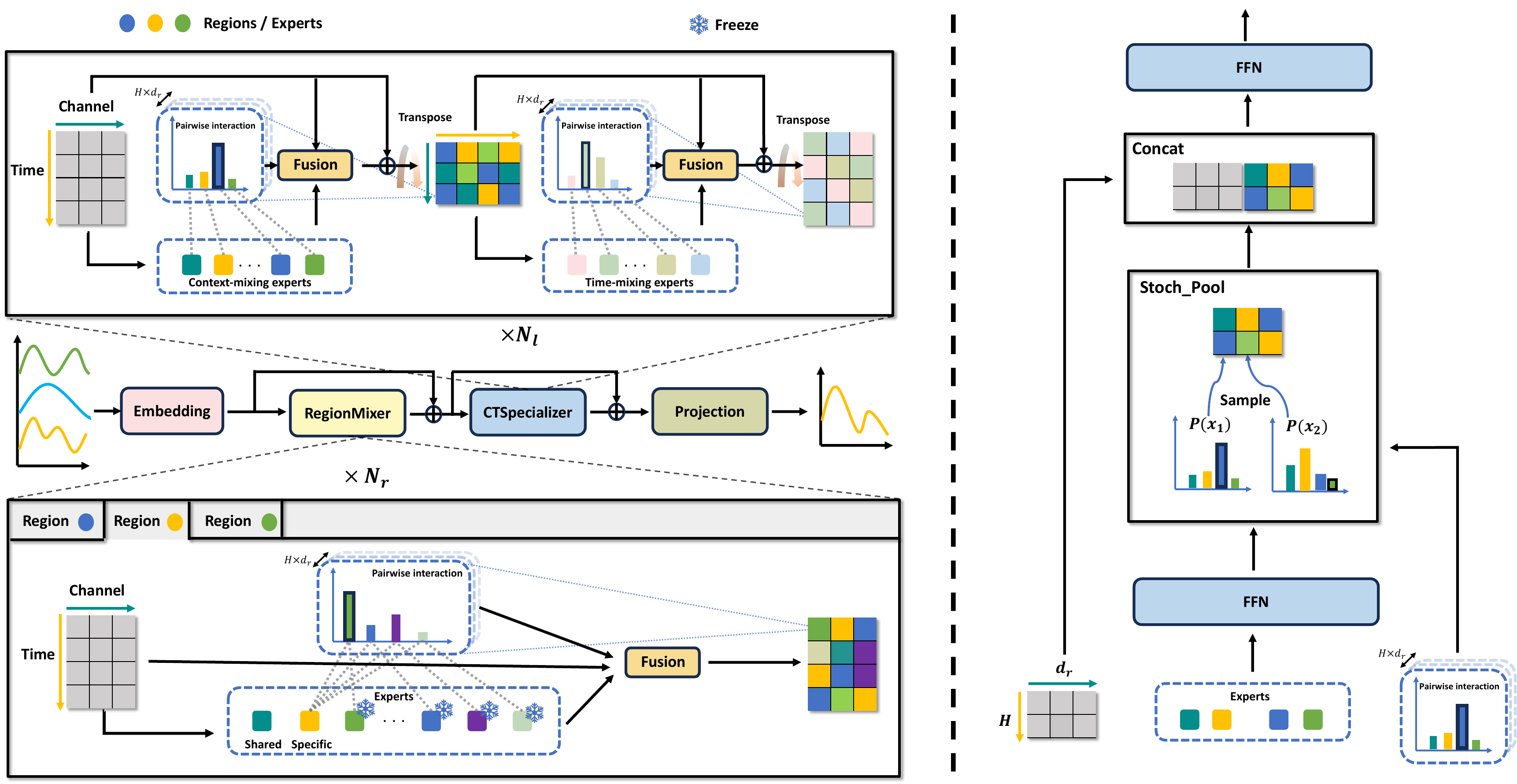}
    \caption{Overall architecture of TriForecaster. The TriForecaster framework is composed of sequential RegionMixer and CTSpecializer layers. RegionMixer layers capture regional information, whereas CTSpecializer layers adaptly address both contextual and temporal variations. Illustrated on the right, these layers are enhanced by a parameter-efficient stochastic fusion module, facilitating effective information aggregation and exploration across various expert components.}
    \label{fig:model}
\end{figure*}

In this section we introduce the detailed design and implementation of our \textit{TriForecaster} framework, as its overview illustrated in Figure \ref{fig:model}. This framework consists of two primary modules:  RegionMixer and Context-Time Specializer (CTSpecializer). The RegionMixer module is designed to integrate region-specific and region-generic information comprehensively across all regions. This integrated information is subsequently processed by the CTSpecializer component, which is designed to effectively capture both contextual and temporal variations. The final electric load forecasts are generated by employing a region-specific projection head on the latent embeddings, which are enriched with a synergistic blend of region-generic, region, contextual, and temporal-specific information.
Following, we introduce these components in detail.

\subsection{RegionMixer}
\label{section:regionmixer}
The goal of RegionMixer is to capture and integrate three types of information: (1) Region-generic information, which encompasses the shared characteristics common to all regions, such as universal dependencies on weather conditions; (2) Region-specific information, which encodes the unique characteristics and factors varying in significance across regions, such as the impact of electricity pricing in industrial hubs; and (3) Pairwise interactions between regions, capturing the relationships where regions with similar industrial structures and geographical locations exhibit analogous behaviors. These types of information are further combined through a stochastic fusion mechanism.

To facilitate the learning of these types of information, prior work~\citep{chen2023mod, ye2023taskexpert, yang2024multi} typically employs task-specific gating networks with Top-K expert selection and auxiliary loss regularization to induce sparsity. However, these approaches can lead to prohibitive computational overhead and risk suboptimal expert specialization due to artificial workload balancing constraints. In contrast, our calculation of pairwise interactions and stochastic fusion mechanism eliminates the need for region-specific gating networks and auxiliary loss regularization, while simultaneously promoting exploration among experts.

As illustrated in Figure \ref{fig:model}, the raw input $\mathbf{X}^{t}$ for region $t$ first goes through an embedding layer $\text{Embedding}(\cdot)$ that projects historical observations with length $L$ to future forecast horizon length $H$ and produces initial latent representation $\mathbf{X}^{t, 0} \in \mathbb{R}^{H \times d_r}$ where $d_r$ is the dimension of the latent space. The latent representation $\mathbf{X}^{t, 0}$ is subsequently processed through $N_r$ stacked RegionMixer layers, each augmented with residual connections~\citep{he2016deep}.

\textbf{Region-generic and specific information extraction.}
For each layer \( l \), the architecture incorporates \( T \) region-specific expert functions \( \{ f^{t,l}: \mathbb{R}^{H \times d_r} \mapsto \mathbb{R}^{H \times d_r} \}_{t=1}^T \) 
, each specialized to encode localized spatiotemporal patterns unique to region \( t \). A shared expert function \( f^{0,l}: \mathbb{R}^{H \times d_r} \mapsto \mathbb{R}^{H \times d_r} \) concurrently models global dynamics common across all regions. Given an input instance \( \mathbf{X}^{t,l}_n \in \mathbb{R}^{H \times d_r} \) from region \( t \) (where \( n \) indexes samples within a batch), the corresponding regional and shared representations are computed as:  
\[
\mathbf{O}^{t,l}_n = f^{t,l}(\mathbf{X}^{t,l}_n), \quad \mathbf{O}^{0,l}_n = f^{0,l}(\mathbf{X}^{t,l}_n),
\]  
both residing in \( \mathbb{R}^{H \times d_r} \). Let \( o^{t,l}_{n,j,k} \in \mathbb{R} \) denote the activation at position \( (j,k) \) of \( \mathbf{O}^{t,l}_n \), and \( f^{t,l}_{j,k} \) represent the functional component generating this activation. It is important to note that \(f\) can be any time series forecasting model. In this study, we specifically employ the TSMixer~\citep{chen2023tsmixer}. The Layer index \( l \) is omitted in subsequent equations where unambiguous.



\textbf{Pairwise interaction extraction}. For region $t$, we construct a 3D probability tensor $\mathbf{P}^t \in \mathbb{R}^{T \times H \times d_r}$ quantifying pairwise expert affinities for each temporal position $j \in \{1,\cdots, H\}$ and latent dimension $k \in \{1,\cdots,d_r\}$. Given a mini-batch of $N$ samples $\{\mathbf{X}^{t}_n\}_{n=1}^N$, the affinity between expert $t$ and expert $i \neq t$ at position $(j,k)$ is computed as:  
\begin{equation}  
d(f^{t}_{j,k}, f^{i}_{j,k}) = \sum_{n=1}^N \left| o^{t}_{n,j,k} - o^{i}_{n,j,k} \right|, 
\end{equation}  
yielding the categorical distribution:  
\begin{equation}  
\mathbf{P}^t_{:,j,k} = \text{Softmax}\left( -\left[ d(f^{t}_{j,k}, f^{i}_{j,k}) \right]_{i \in \{0,\ldots,T\}\setminus\{t\}} \right),
\end{equation}  
This procedure of calculating pairwise affinities eliminates region-specific gating networks by directly measuring component-wise expert similarity through batch statistics. 

\textbf{Stochastic fusion mechanism}. We utilize a parameter-efficient fusion mechanism through component-wise stochastic fusion~\citep{han2024softs}. Let $\mathbf{O}_{\text{cat}} = \{\mathbf{O}^i\}_{i \in \{0,\ldots,T\}\setminus\{t\}}\in \mathbb{R}^{T \times H \times d_r}$ be the concatenated outputs of experts. The Fusion module:
\begin{equation}  
\mathbf{X}^t = \text{Fusion}(\{\mathbf{O}^t, \mathbf{O}^0\}, \mathbf{O}_{\text{cat}}, 
\mathbf{P}^t) \in \mathbb{R}^{H \times d_r},
\end{equation} 
synthesizes regional ($\mathbf{O}^t$), global ($\mathbf{O}^0$), and pairwise ($\mathbf{O}_{\text{pair}}$) information through:  
\begin{equation}  
\begin{aligned}  
\mathbf{O}_{\text{cat}} &= \text{FFN}_1\left( \mathbf{O}_{\text{cat}} \right) \in \mathbb{R}^{T \times H \times d_r} \\  
\mathbf{O}_{\text{pair}} &= \text{Stoch\_Pool}(\mathbf{O}_{\text{cat}}, \mathbf{P}^t)\\  
\mathbf{X}^t &= \text{FFN}_2\left( \text{Concat}(\mathbf{O}^t, \mathbf{O}^0, \mathbf{O}_{\text{pair}}) \right) \in \mathbb{R}^{H \times d_r},  
\end{aligned}  
\end{equation}  
where $\text{FFN}_1: \mathbb{R}^{d_r} \to \mathbb{R}^{d_r}$ and $\text{FFN}_2: \mathbb{R}^{3d_r} \to \mathbb{R}^{d_r}$ are feed-forward networks, Concat operator concatenate the inputs on the latent dimension, and $\text{Stoch\_Pool}(\mathbf{O}_{\text{cat}}, \mathbf{P}^t) \in \mathbb{R}^{H \times d_r}$ implements a stochastic pooling mechanism defined as:
\begin{equation}
\begin{aligned}
(o_{\text{pair}})_{j, k} = (o_{\text{cat}})_{e, j, k} \quad \text{where} \; e \sim \text{Categorical}(\mathbf{P}^t_{:, j, k}).
\end{aligned}
\end{equation}

This module improves the expressiveness of RegionMixer by allowing each temporal position and latent dimension to select freely among experts. At the same time, by utilizing the stochastic pooling, it encourages the exploration over experts while removing the need for auxiliary loss.

\subsection{CTSpecializer}
\label{section:CTSpecializer}
Following the stack of \(N_r\) RegionMixer layers, the representation \(\mathbf{X}^{t, N_r} \in \mathbb{R}^{H \times d_r}\) is obtained, capturing both region-specific and general information for region \(t\). The aim of the CTSpecializer is to further delineate the contextual and temporal variations through the use of Context-specializing MoE (ContextMoE) and Time-specializing MoE (TimeMoE), respectively. In the Context-specializing MoE, each expert can specialize in different contextual types, whereas in the Time-specializing MoE, each expert can focus on varying spans of the target forecast horizon. These specialized experts are further integrated using the stochastic fusion mechanism, allowing each temporal position and latent dimension to dynamically select among experts specializing in different contexts and time spans throughout the target forecast horizon.

Let \(\mathbf{X}^{t, 0} = \mathbf{X}^{t, N_r}\) denote the initial latent representation for a stack of \(N_{l}\) CTSpecializer layers. As illustrated in Figure \ref{fig:model}, the process for any layer \(l \in \{1,\cdots, N_{l}\}\) is described by:
\begin{equation}
\begin{aligned}
\mathbf{X}^{t, l} = \text{CTSpecializer}(\mathbf{X}^{t, l-1}) + \mathbf{X}^{t, l-1}.
\end{aligned}
\end{equation}
Specifically, CTSpecializer comprises two key components:
\begin{itemize}
    \item \textbf{Context-specializing MoE}. This component captures contextual variations between different instances, employing a set of experts \(\{g^{i, l}:  \mathbb{R}^{H \times d_r} \mapsto \mathbb{R}^{H \times d_r}\}^{N_c}_{i=1}\). These experts, implemented as multi-layer perceptrons (MLPs), operate on the latent dimension for each timestamp. For input \(\mathbf{X}^{t, l-1}\), the 3D probability tensor \(\mathbf{P}^{l} \in \mathbb{R}^{T \times H \times d_r}\) representing the pairwise interaction between the input and experts, is constructed directly from expert activations~\citep{han2024softs}, as follows:
    \begin{equation}  
    \mathbf{P}^{l}_{:,j,k} = \text{Softmax}\left( \left[g^{i, l}_{j, k} (\mathbf{X}^{t, l-1})\right]^{N_c}_{i=1} \right).
    \end{equation}  
    The Fusion module, as defined in Section \ref{section:regionmixer}, is applied to fuse the concatenated outputs of experts:
    \begin{equation}
    \mathbf{O}^l_{\text{cat}} = \{g^{i, l}(\mathbf{X}^{t,l})\}^{N_c}_{i=1} \in \mathbb{R}^{N_c \times H \times d_r},
    \end{equation} 
    with the input $\mathbf{X}^{t, l-1}$:
    \begin{equation}  
    \mathbf{\mathbf{X}}^{t, l} = \text{Fusion}\left(\mathbf{X}^{t, l-1},\mathbf{O}_{\text{cat}}, \mathbf{P}^{l}\right) \in \mathbb{R}^{H \times d_r}.
    \end{equation}  
    \item \textbf{Time-specializing MoE}. This component addresses temporal variations across instances. The output from context-specializing stage, \(\mathbf{X}^{t, l} \in \mathbb{R}^{H \times d_r}\), is transposed to \(\mathbf{X}^{t, l} \in \mathbb{R}^{d_r \times H}\) before being processed by a set of MLP experts \(\{h^{i, l}:  \mathbb{R}^{d_r \times H} \mapsto \mathbb{R}^{d_r \times H}\}^{N_t}_{i=1}\). A corresponding 3D probability tensor \(\mathbf{P}^{l} \in \mathbb{R}^{T \times d_r \times H}\) is similarly constructed from the resulting activations and the Fusion module is applied to fuse the outputs of experts with the input $\mathbf{X}^{t, l}$. The result is transposed back to \(\mathbf{X}^{t, l} \in \mathbb{R}^{H \times d_r}\).
\end{itemize}

In traditional manual forecasting methods, a historical interval of length $H$ is initially selected based on its contextual similarity to the future forecast horizon. The forecast is then constructed by examining the differences in covariate values between this historical interval and the intended forecast horizon. Drawing inspiration from this approach, we encourage contextually similar samples to be processed by similar context-specializing experts. This is achieved through the integration of a contrastive loss, which encourages the model to differentiate between varying contexts by reinforcing similarities and distinctions among the samples. Specifically, for each raw training sample \(\mathbf{X}_n^t\) from region \(t\), a positive sample \(\overline{\mathbf{X}}_n^t\) and a set of negative samples \(\{\widetilde{\mathbf{X}}_i^t\}_{i=1}^{N_n}\) are generated. These samples are differentiated based on the Euclidean distances computed across salient covariates such as weather conditions, electricity pricing, and date. In any CTSpecializer layer \(l \in \{1, \cdots, N_l\}\), the context-specialized outputs for the primary training sample, its positive, and negative counterparts are denoted as \(\mathbf{X}_n^{t, l}\), \(\overline{\mathbf{X}}_n^{t, l}\), and \(\{\widetilde{\mathbf{X}}_i^{t, l}\}_{i=1}^{N_n}\), respectively. The contrastive loss for the training sample \(n\) is expressed as:

\begin{equation}\label{eq:contrastive:loss} 
l^{t}_{n} = -\frac{\exp\left(\text{sim}(\mathbf{X}_n^{t, l}, \overline{\mathbf{X}}_n^{t, l}) / \tau\right)}{\exp\left(\text{sim}(\mathbf{X}_n^{t, l}, \overline{\mathbf{X}}_n^{t, l}) / \tau\right) + \sum^{N_n}_{i=1}\exp\left(\text{sim}(\mathbf{X}_n^{t, l}, \widetilde{\mathbf{X}}_n^{t, l}) / \tau\right)},
\end{equation}  
where $\tau$ is the temperature parameter~\citep{chen2020simple} and the similarity $\text{sim}(\mathbf{U}, \mathbf{V})$ between two matrices $\mathbf{U}, \mathbf{V} \in \mathbb{R}^{H \times d_r}$ is calculated by first flatten matrices into vectors and apply cosine similarities between two vectors using dot product and $l_2$ norm,  formulated as:
\begin{equation} 
\begin{aligned}
\mathbf{u} &= \text{Flatten}(\mathbf{U})\in \mathbb{R}^{(H \times d_r)},\\
\mathbf{v} &= \text{Flatten}(\mathbf{V}) \in \mathbb{R}^{(H \times d_r)},\\
\text{sim}(\mathbf{\mathbf{u}}, \mathbf{v}) &= \frac{\mathbf{u} \cdot \mathbf{v}}{\|\mathbf{u}\|\|\mathbf{v}\|}.
\end{aligned}
\end{equation}  

\subsection{Loss function} For a region \(t\) with a mini-batch \(\{\mathbf{X}^t_n\}_{n=1}^N\), after processing through \(N_r\) RegionMixer layers and \(N_l\) CTSpecializer layers, predictions are generated using a region-specific projection head \(\phi_t: \mathbb{R}^{H \times d_r} \rightarrow \mathbb{R}^{H \times 1}\):
\begin{equation} 
\mathbf{\hat{y}}_n = \phi_t(\mathbf{X}^{t, N_l}_n).
\end{equation}  
The complete loss function integrates both the mean squared error (MSE) loss and the contrastive loss, defined as follows:
\begin{equation} 
\mathcal{L} = \frac{1}{N}\sum_{t=1}^{T}\sum_{i=1}^{N} (\mathbf{\hat{y}}_i - \mathbf{y}_i)^2 + \alpha \cdot l^t_{i}, 
\end{equation} 
where the contrastive loss $l^t_i$ is defined in Eq.~\eqref{eq:contrastive:loss}, and \(\alpha\) denotes the regularization coefficient balancing the contributions between the MSE and contrastive losses.
\section{Experiments}
In this section, we aim to validate the effectiveness of our proposed method, TriForecaster, by addressing the following research questions: (\textbf{RQ1}) Can TriForecaser outperform SOTA MRELF methods when applied to real-world datasets characterized by varying levels of granularity?
(\textbf{RQ2}) What is the contribution of each component of the TriForecaster in tackling the MRELF problem?

\begin{table*}[t]
\caption{The final comparison performance on test sets averaged over five random seeds. The bold values are the best results.
}
\label{table:comparison_performance}
\normalsize
\centering
\renewcommand{\arraystretch}{0.8}
\begin{tabular}{c| cc| cc| cc | cc}
\toprule
 & \multicolumn{2}{c|}{EPC} & \multicolumn{2}{c|}{CEESC} & \multicolumn{2}{c|}{City-load} & \multicolumn{2}{c}{Bus-load} \\
\multirow{-2}{*}{Method} & MSE & MAE & MSE & MAE & MSE & MAE & MSE & MAE\\
\midrule
STL & 0.3816 & 0.4761 & 0.1415  & 0.3086 & 0.1486 & 0.2813 & 0.3201 & 0.3913  \\
MTL &  0.1222 &  0.2579 &  0.1208  &  0.2801 &  0.1021 &  0.2273 & \underline{0.2327} & \underline{0.3543}  \\
 MLoRE & \underline{0.1008} & \underline{0.2479} & \underline{0.0405}  & \underline{0.1383} & 0.1034  & 0.2316 & 0.2630 & 0.3768  \\
 GCN-LSTM & 0.4342 & 0.5183 & 0.1905 & 0.3376 & \underline{0.0912} & \underline{0.2243}  & 0.2585 & 0.3828  \\
 \midrule
 TriForecaster(ours) & \textbf{0.0795} &	\textbf{0.2092}	&  \textbf{0.0267} &	\textbf{0.1232} & \textbf{0.0848}	& \textbf{0.2070}  & \textbf{0.1686} & \textbf{0.3009}  \\
\bottomrule
\end{tabular}

\end{table*}

\subsection{Experimental Settings}
\textbf{Datasets.}\hspace{0.5mm} We evaluate our method on four multiple-region load forecasting datasets with different levels of granularity:
(1) \textbf{EPC}\footnote{\url{https://www.kaggle.com/datasets/fedesoriano/electric-power-consumption}} ~\citep{power-consumption-prediction} dataset comprises electric power consumption alongside nine related meteorological variables. The data spans one year and is collected from three different zone stations in Tétouan city, Morocco. Observations are recorded every 10 minutes. For our experiments, we use a look-back window of 504 time steps and a forecast horizon of 216 time steps (36 hours);
(2) \textbf{CEESC} ~\citep{electrician-competition} dataset originates from the 9th China Electrical Engineering Society Cup National Undergraduate Electrical Mathematical Modeling Competition. It contains electricity load data recorded at 15-minute intervals, alongside five meteorological variables, spanning from January 1, 2012, to January 10, 2015. The data is collected from two specific areas. For our experiments, we utilize a look-back window of 336 time steps and a forecast horizon of 144 time steps (36 hours);
(3) \textbf{City-load} consists of real-world electric load records for 17 different cities in a province located on the eastern coast of China. The data covers the period from 2022 to 2024, with observations collected at 15-minute intervals. Additionally, the dataset includes 18 meteorological variables relevant to the regions. We apply a look-back window of 400 time steps and a forecast horizon of 144 time steps;
(4) \textbf{Bus-load} provides real-world electric load data for six buses operating in different areas of a province on the eastern coast of China. The data is collected from January 2020 to July 2022 at 15-minute intervals and is accompanied by 18 corresponding meteorological variables. For this dataset, we utilize a look-back window of 400 time steps and a forecast horizon of 144 time steps.

\textbf{Baselines.}  We compare our method with multiple baselines, including: (1) \textbf{STL}:  A  na\"{\i}ve approach involves training models independently for each region; (2) \textbf{MTL}: a multi-task learning baseline utilizing a unified backbone model with distinct prediction heads for each region; (3) \textbf{MLoRE}~\citep{MLoRE}: a SOTA method for multi-task learning in the field of vision using Mixture of Low-Rank Experts adapted to MRELF; (4) \textbf{GCN-LSTM}~\citep{mansoor2024graph}: a SOTA MRELF method that employs Graph Convolutional Networks (GCN)~\citep{kipf2016semi} for spatial and LSTM~\citep{hochreiter1997long} for temporal feature extraction. For all baselines, We replace the backbone model with TSMixer~\citep{chen2023tsmixer}.

\textbf{Implementation Details.}
For all datasets, we use the second last month of data as the validation set and the last month as the test set. In training phase, we use Adam~\citep{kingma2014adam} with fixed learning rate of 0.001 and batch size 64 to optimize all parameters w.r.t $l_2$ (Mean Squared Loss) loss. We perform hyperparameter tuning on the validation set for all baselines, with further details provided in the Appendix \ref{appendix:backbone_baseline_details}. All experimental results are the average of the five independent trials with different random seeds. We also conduct hyperparameter sensitivity analyses, with more details in Appendix \ref{appendix:hyper_sen}.


\subsection{Performance Comparison (RQ1)}

The results of the performance comparison are presented in Table \ref{table:comparison_performance}, with the corresponding standard deviations provided in Table \ref{table:std}. We employ mean squared error (MSE) and mean absolute error (MAE) as our evaluation metrics, both averaged over test datasets derived from five random seed iterations. Key insights include:  
(1) TriForecaster achieves a 22.4\% MSE reduction and 12.3\% MAE reduction compared to the second-best method across all datasets, with notable improvements on the CEESC (34.1\% MSE reduction) and Bus-load (27.5\% MSE reduction) datasets; (2) STL underperforms due to overfitting on noisy, limited data, highlighting the necessity of leveraging cross-regional shared information; (3) TriForecaster excels on fine-grained datasets (e.g., EPC, CEESC, Bus-load) with heterogeneous regions (e.g., stations, buses), where methods like GCN-LSTM and MTL struggles. Furthermore, TriForecaster’s superior performance over MLoRE across all datasets showcases its effective specialization capabilities across regional, contextual, and temporal dimensions. We note here that MLoRE is orthogonal to our method which can be easily integrated into TriForecaster.
\subsection{Ablation Studies (RQ2)}
We conduct ablation studies to validate the effectiveness of RegionMixer, Context-specializing MoE, and Time-specializing MoE. To evaluate their contributions, we sequentially remove (w/o) each component: (1) \textbf{w/o RegionMixer} removes RegionMixer layers, the embedding is directly feed into the CTSpecializer layers; (2) \textbf{w/o ContextMoE} removes Context-specializing MoE modules from the CTSpecializer layers; (3) \textbf{w/o TimeMoE} removes the Time-specializing MoE modules from the CTSpecializer layers; (4) \textbf{FriForecaster} is the standard TriForecaster that incorporates both RegionMixer layers and CTSpecializer layers.

The results are presented in Table \ref{table:ablation_study}. We observe that removing any module significantly degrades forecast performance. This finding aligns with our observations of regional, contextual, and temporal variations, highlighting the importance and effectiveness of each module within our framework.

\begin{table}[t]
\caption{The ablation study results using EPC dataset. The bold values are the best results.}
\label{table:ablation_study}
\renewcommand\arraystretch{0.9}
\tiny 
\centering
\resizebox{0.7\columnwidth}{!}{
\begin{tabular}{c|cc}
\toprule
 & \multicolumn{2}{c}{EPC}\\
 \multirow{-2}{*}{Method} & MSE & MAE\\
\midrule
 w/o RegionMixer & 0.0925 & 0.2288\\
 w/o TimeMoE & 0.0981 & 0.2408 \\
 w/o ContextMoE & 0.0966  & 0.2355  \\
 TriForecaster & \textbf{0.0795} & \textbf{0.2092 }  \\
\bottomrule
\end{tabular}
}

\end{table}

\begin{figure}[!t]
    \centering
    \includegraphics[width=\linewidth]{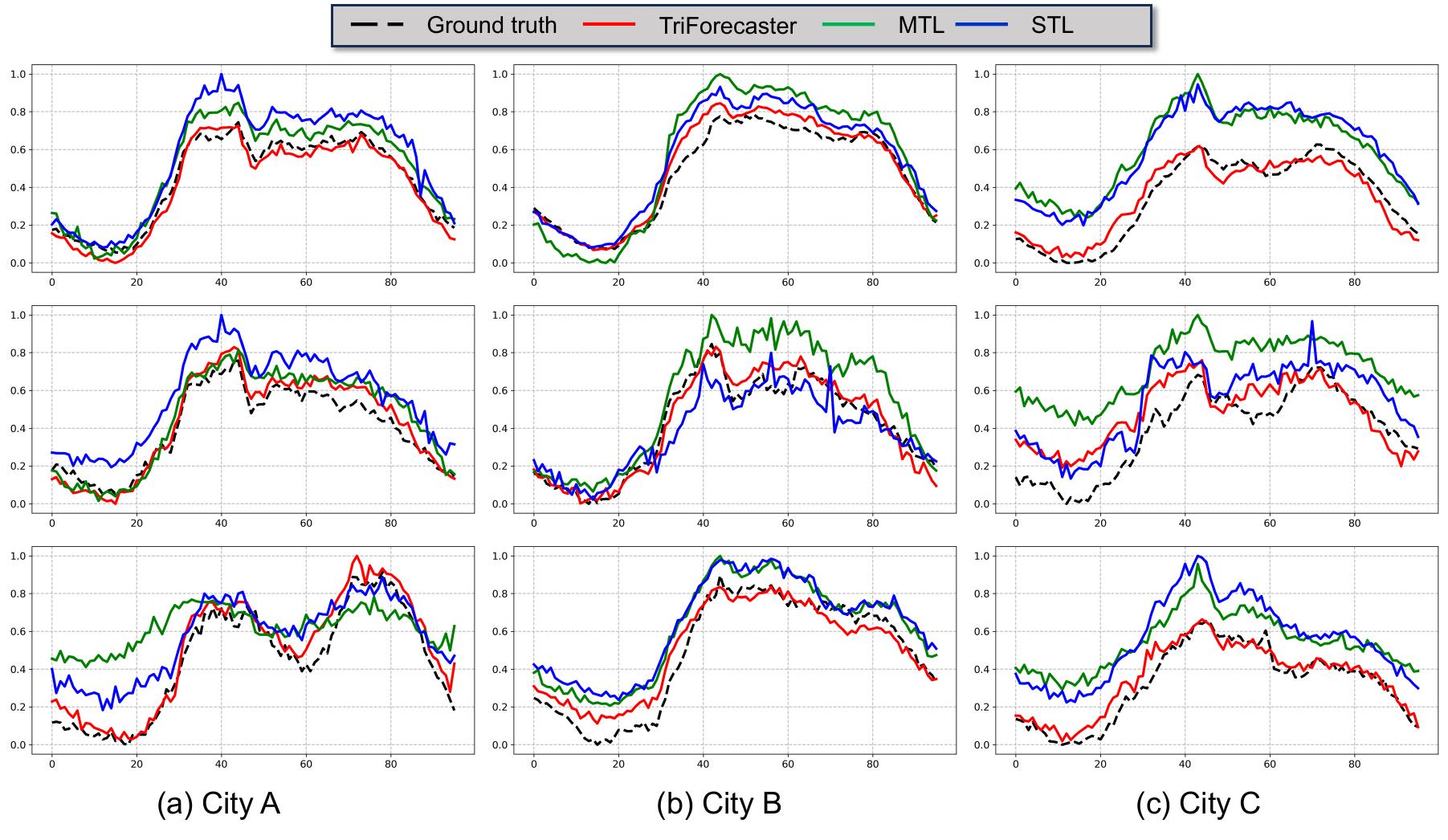}
    \caption{The predictions of TriForecaster, MTL, and STL on three cities under varying contexts using the City-load dataset.}
    \label{fig:case-study}
\end{figure}

\subsection{Case Studies}

In this section, we examine the forecasting capability of TriForecaster in the presence of regional, contextual, and temporal variations through a series of case studies. Figure \ref{fig:case-study} illustrates the predictions made by TriForecaster, MTL, and STL for three cities within the City-load dataset. These predictions are evaluated under varying contexts, where substantial differences in shapes and scales are evident due to the aforementioned variations. In all scenarios, STL forecasts exhibit a significant number of abnormal spikes and fluctuations, indicating a considerable degree of overfitting. Conversely, MTL provides forecasts that are similar across most cases, failing to adequately capture the unique characteristics inherent in the presence of variations. In comparison to these baseline methods, TriForecaster delivers robust forecasts with minimal fluctuations and abnormal spikes. It consistently outperforms the baseline methods across different cities, contexts, and time spans of the forecast horizons in all scenarios.

\section{Deployment}\label{sec:deployment}


Since June 2024, TriForecaster has been successfully implemented to predict daily electricity loads at the municipal level for 17 cities, collectively serving a population exceeding 110 million and managing a daily electricity consumption of over 100 gigawatt-hours. The predictions are utilized by experts in grid company to make daily trading, planning, and scheduling decisions. Our system consistently outperforms the previous model, which employed STL, by a margin of 1.6\% in terms of average prediction accuracy. This enhancement translates to an approximate saving of 2 million kilowatt-hours daily, which is anticipated to result in significant cost savings of around 300 thousand U.S. dollars per day.

TriForecaster is seamlessly integrated into the eForecaster platform \citep{zhu2023eforecaster}. For each city, raw electricity data, weather data, and supplementary data (e.g., factory production plans, electricity pricing) are initially processed and combined in the input module. These datasets are then refined in the preprocessing module using a combination of robust time series decomposition, anomaly detection, and missing value imputation methods \citep{wen2021robustperiod, RobustSTL_wen2018robuststl, FastRobustSTL_wen2020}. Subsequently, future covariates, such as date, holiday, and pricing information, are generated. To enhance efficiency, multiprocessing is employed across cities during these steps. Following this, all datasets are fed into TriForecaster to produce preliminary forecasts for each city. These preliminary forecasts are further refined by the postprocessing module, which integrates domain knowledge with model forecasts by directly influencing the model outputs for each city.

\section{Conclusion and Future Work}
In conclusion, our proposed TriForecaster framework provoides an effective solution to the complex challenges of Multi-region Electric Load Forecasting problem (MRELF) through the integration of RegionMixer and CTSpecializer layers. These layers enable effective collaboration and specialization across regional, contextual, and temporal dimensions. Our versatile framework has broad applicability, making it suitable for addressing a wide range of real-world MRELF problems across different levels of granularity. In the future, we plan to deploy our TriForecaster based on our general eForecaster platform in more regions. 

\section{GenAI Usage Disclosure}

AI tools (GPT-4o) are only used for automated grammar corrections. 


\newpage
\newpage
\bibliographystyle{ACM-Reference-Format}
\bibliography{reference}


\begin{thebibliography}{45}


\ifx \showCODEN    \undefined \def \showCODEN     #1{\unskip}     \fi
\ifx \showISBNx    \undefined \def \showISBNx     #1{\unskip}     \fi
\ifx \showISBNxiii \undefined \def \showISBNxiii  #1{\unskip}     \fi
\ifx \showISSN     \undefined \def \showISSN      #1{\unskip}     \fi
\ifx \showLCCN     \undefined \def \showLCCN      #1{\unskip}     \fi
\ifx \shownote     \undefined \def \shownote      #1{#1}          \fi
\ifx \showarticletitle \undefined \def \showarticletitle #1{#1}   \fi
\ifx \showURL      \undefined \def \showURL       {\relax}        \fi
\providecommand\bibfield[2]{#2}
\providecommand\bibinfo[2]{#2}
\providecommand\natexlab[1]{#1}
\providecommand\showeprint[2][]{arXiv:#2}

\bibitem[Abumohsen et~al\mbox{.}(2023)]%
        {abumohsen2023electrical}
\bibfield{author}{\bibinfo{person}{Mobarak Abumohsen}, \bibinfo{person}{Amani~Yousef Owda}, {and} \bibinfo{person}{Majdi Owda}.} \bibinfo{year}{2023}\natexlab{}.
\newblock \showarticletitle{Electrical load forecasting using LSTM, GRU, and RNN algorithms}.
\newblock \bibinfo{journal}{\emph{Energies}} \bibinfo{volume}{16}, \bibinfo{number}{5} (\bibinfo{year}{2023}), \bibinfo{pages}{2283}.
\newblock


\bibitem[Asiri et~al\mbox{.}(2024)]%
        {asiri2024short}
\bibfield{author}{\bibinfo{person}{Mashael~M Asiri}, \bibinfo{person}{Ghadah Aldehim}, \bibinfo{person}{Faiz~Abdullah Alotaibi}, \bibinfo{person}{Mrim~M Alnfiai}, \bibinfo{person}{Mohammed Assiri}, {and} \bibinfo{person}{Ahmed Mahmud}.} \bibinfo{year}{2024}\natexlab{}.
\newblock \showarticletitle{Short-term load forecasting in smart grids using hybrid deep learning}.
\newblock \bibinfo{journal}{\emph{IEEE Access}}  \bibinfo{volume}{12} (\bibinfo{year}{2024}), \bibinfo{pages}{23504--23513}.
\newblock


\bibitem[Bai et~al\mbox{.}(2018)]%
        {bai2018empirical}
\bibfield{author}{\bibinfo{person}{Shaojie Bai}, \bibinfo{person}{J.~Zico Kolter}, {and} \bibinfo{person}{Vladlen Koltun}.} \bibinfo{year}{2018}\natexlab{}.
\newblock \bibinfo{title}{An Empirical Evaluation of Generic Convolutional and Recurrent Networks for Sequence Modeling}.
\newblock
\showeprint[arxiv]{1803.01271}


\bibitem[Cai et~al\mbox{.}(2022)]%
        {cai2022short}
\bibfield{author}{\bibinfo{person}{Changchun Cai}, \bibinfo{person}{Yuanjia Li}, \bibinfo{person}{Zhenghua Su}, \bibinfo{person}{Tianqi Zhu}, {and} \bibinfo{person}{Yaoyao He}.} \bibinfo{year}{2022}\natexlab{}.
\newblock \showarticletitle{Short-term electrical load forecasting based on VMD and GRU-TCN hybrid network}.
\newblock \bibinfo{journal}{\emph{Applied Sciences}} \bibinfo{volume}{12}, \bibinfo{number}{13} (\bibinfo{year}{2022}), \bibinfo{pages}{6647}.
\newblock


\bibitem[Chen et~al\mbox{.}(2023c)]%
        {chen2023research}
\bibfield{author}{\bibinfo{person}{Houhe Chen}, \bibinfo{person}{Mingyang Zhu}, \bibinfo{person}{Xiao Hu}, \bibinfo{person}{Jiarui Wang}, \bibinfo{person}{Yong Sun}, {and} \bibinfo{person}{Jinduo Yang}.} \bibinfo{year}{2023}\natexlab{c}.
\newblock \showarticletitle{Research on short-term load forecasting of new-type power system based on GCN-LSTM considering multiple influencing factors}.
\newblock \bibinfo{journal}{\emph{Energy Reports}}  \bibinfo{volume}{9} (\bibinfo{year}{2023}), \bibinfo{pages}{1022--1031}.
\newblock


\bibitem[Chen et~al\mbox{.}(2023a)]%
        {chen2023tsmixer}
\bibfield{author}{\bibinfo{person}{Si-An Chen}, \bibinfo{person}{Chun-Liang Li}, \bibinfo{person}{Nate Yoder}, \bibinfo{person}{Sercan~O Arik}, {and} \bibinfo{person}{Tomas Pfister}.} \bibinfo{year}{2023}\natexlab{a}.
\newblock \showarticletitle{Tsmixer: An all-mlp architecture for time series forecasting}.
\newblock \bibinfo{journal}{\emph{arXiv preprint arXiv:2303.06053}} (\bibinfo{year}{2023}).
\newblock


\bibitem[Chen and Guestrin(2016)]%
        {chen2016xgboost}
\bibfield{author}{\bibinfo{person}{Tianqi Chen} {and} \bibinfo{person}{Carlos Guestrin}.} \bibinfo{year}{2016}\natexlab{}.
\newblock \showarticletitle{Xgboost: A scalable tree boosting system}. In \bibinfo{booktitle}{\emph{Proceedings of the 22nd ACM SIGKDD international conference on knowledge discovery and data mining}}. \bibinfo{pages}{785--794}.
\newblock


\bibitem[Chen et~al\mbox{.}(2020a)]%
        {chen2020simple}
\bibfield{author}{\bibinfo{person}{Ting Chen}, \bibinfo{person}{Simon Kornblith}, \bibinfo{person}{Mohammad Norouzi}, {and} \bibinfo{person}{Geoffrey Hinton}.} \bibinfo{year}{2020}\natexlab{a}.
\newblock \showarticletitle{A simple framework for contrastive learning of visual representations}. In \bibinfo{booktitle}{\emph{International conference on machine learning}}. PMLR, \bibinfo{pages}{1597--1607}.
\newblock


\bibitem[Chen et~al\mbox{.}(2022)]%
        {chen2022learning}
\bibfield{author}{\bibinfo{person}{Weiqi Chen}, \bibinfo{person}{Wenwei Wang}, \bibinfo{person}{Bingqing Peng}, \bibinfo{person}{Qingsong Wen}, \bibinfo{person}{Tian Zhou}, {and} \bibinfo{person}{Liang Sun}.} \bibinfo{year}{2022}\natexlab{}.
\newblock \showarticletitle{Learning to Rotate: Quaternion Transformer for Complicated Periodical Time Series Forecasting}. In \bibinfo{booktitle}{\emph{Proceedings of the 28th ACM SIGKDD Conference on Knowledge Discovery and Data Mining}}. \bibinfo{pages}{146--156}.
\newblock


\bibitem[Chen et~al\mbox{.}(2020b)]%
        {chen2020just}
\bibfield{author}{\bibinfo{person}{Zhao Chen}, \bibinfo{person}{Jiquan Ngiam}, \bibinfo{person}{Yanping Huang}, \bibinfo{person}{Thang Luong}, \bibinfo{person}{Henrik Kretzschmar}, \bibinfo{person}{Yuning Chai}, {and} \bibinfo{person}{Dragomir Anguelov}.} \bibinfo{year}{2020}\natexlab{b}.
\newblock \showarticletitle{Just pick a sign: Optimizing deep multitask models with gradient sign dropout}.
\newblock \bibinfo{journal}{\emph{Advances in Neural Information Processing Systems}}  \bibinfo{volume}{33} (\bibinfo{year}{2020}), \bibinfo{pages}{2039--2050}.
\newblock


\bibitem[Chen et~al\mbox{.}(2023b)]%
        {chen2023mod}
\bibfield{author}{\bibinfo{person}{Zitian Chen}, \bibinfo{person}{Yikang Shen}, \bibinfo{person}{Mingyu Ding}, \bibinfo{person}{Zhenfang Chen}, \bibinfo{person}{Hengshuang Zhao}, \bibinfo{person}{Erik~G Learned-Miller}, {and} \bibinfo{person}{Chuang Gan}.} \bibinfo{year}{2023}\natexlab{b}.
\newblock \showarticletitle{Mod-squad: Designing mixtures of experts as modular multi-task learners}. In \bibinfo{booktitle}{\emph{Proceedings of the IEEE/CVF Conference on Computer Vision and Pattern Recognition}}. \bibinfo{pages}{11828--11837}.
\newblock


\bibitem[Chung et~al\mbox{.}(2014)]%
        {chung2014empirical}
\bibfield{author}{\bibinfo{person}{Junyoung Chung}, \bibinfo{person}{Caglar Gulcehre}, \bibinfo{person}{KyungHyun Cho}, {and} \bibinfo{person}{Yoshua Bengio}.} \bibinfo{year}{2014}\natexlab{}.
\newblock \showarticletitle{Empirical evaluation of gated recurrent neural networks on sequence modeling}.
\newblock \bibinfo{journal}{\emph{arXiv preprint arXiv:1412.3555}} (\bibinfo{year}{2014}).
\newblock


\bibitem[Fan et~al\mbox{.}(2010)]%
        {Fan:MREFL:2010}
\bibfield{author}{\bibinfo{person}{Shu Fan}, \bibinfo{person}{Kittipong Methaprayoon}, {and} \bibinfo{person}{Wei-Jen Lee}.} \bibinfo{year}{2010}\natexlab{}.
\newblock \showarticletitle{Multi-region load forecasting considering alternative meteorological predictions}. In \bibinfo{booktitle}{\emph{IEEE PES General Meeting}}. \bibinfo{pages}{1--7}.
\newblock
\href{https://doi.org/10.1109/PES.2010.5589869}{doi:\nolinkurl{10.1109/PES.2010.5589869}}


\bibitem[Fida et~al\mbox{.}(2024)]%
        {fida2024comprehensive}
\bibfield{author}{\bibinfo{person}{Kinza Fida}, \bibinfo{person}{Usman Abbasi}, \bibinfo{person}{Muhammad Adnan}, \bibinfo{person}{Sajid Iqbal}, {and} \bibinfo{person}{Salah Eldeen~Gasim Mohamed}.} \bibinfo{year}{2024}\natexlab{}.
\newblock \showarticletitle{A comprehensive survey on load forecasting hybrid models: Navigating the Futuristic demand response patterns through experts and intelligent systems}.
\newblock \bibinfo{journal}{\emph{Results in Engineering}} (\bibinfo{year}{2024}), \bibinfo{pages}{102773}.
\newblock


\bibitem[Fiot and Dinuzzo(2016)]%
        {fiot2016electricity}
\bibfield{author}{\bibinfo{person}{Jean-Baptiste Fiot} {and} \bibinfo{person}{Francesco Dinuzzo}.} \bibinfo{year}{2016}\natexlab{}.
\newblock \showarticletitle{Electricity demand forecasting by multi-task learning}.
\newblock \bibinfo{journal}{\emph{IEEE Transactions on Smart Grid}} \bibinfo{volume}{9}, \bibinfo{number}{2} (\bibinfo{year}{2016}), \bibinfo{pages}{544--551}.
\newblock


\bibitem[Han et~al\mbox{.}(2024)]%
        {han2024softs}
\bibfield{author}{\bibinfo{person}{Lu Han}, \bibinfo{person}{Xu-Yang Chen}, \bibinfo{person}{Han-Jia Ye}, {and} \bibinfo{person}{De-Chuan Zhan}.} \bibinfo{year}{2024}\natexlab{}.
\newblock \showarticletitle{SOFTS: Efficient Multivariate Time Series Forecasting with Series-Core Fusion}.
\newblock \bibinfo{journal}{\emph{arXiv preprint arXiv:2404.14197}} (\bibinfo{year}{2024}).
\newblock


\bibitem[He et~al\mbox{.}(2016)]%
        {he2016deep}
\bibfield{author}{\bibinfo{person}{Kaiming He}, \bibinfo{person}{Xiangyu Zhang}, \bibinfo{person}{Shaoqing Ren}, {and} \bibinfo{person}{Jian Sun}.} \bibinfo{year}{2016}\natexlab{}.
\newblock \showarticletitle{Deep residual learning for image recognition}. In \bibinfo{booktitle}{\emph{Proceedings of the IEEE conference on computer vision and pattern recognition}}. \bibinfo{pages}{770--778}.
\newblock


\bibitem[Hochreiter(1997)]%
        {hochreiter1997long}
\bibfield{author}{\bibinfo{person}{S Hochreiter}.} \bibinfo{year}{1997}\natexlab{}.
\newblock \showarticletitle{Long Short-term Memory}.
\newblock \bibinfo{journal}{\emph{Neural Computation MIT-Press}} (\bibinfo{year}{1997}).
\newblock


\bibitem[Imani(2021)]%
        {imani2021electrical}
\bibfield{author}{\bibinfo{person}{Maryam Imani}.} \bibinfo{year}{2021}\natexlab{}.
\newblock \showarticletitle{Electrical load-temperature CNN for residential load forecasting}.
\newblock \bibinfo{journal}{\emph{Energy}}  \bibinfo{volume}{227} (\bibinfo{year}{2021}), \bibinfo{pages}{120480}.
\newblock


\bibitem[Jiang et~al\mbox{.}(2021)]%
        {jiang2021hybrid}
\bibfield{author}{\bibinfo{person}{Lianjie Jiang}, \bibinfo{person}{Xinli Wang}, \bibinfo{person}{Wei Li}, \bibinfo{person}{Lei Wang}, \bibinfo{person}{Xiaohong Yin}, {and} \bibinfo{person}{Lei Jia}.} \bibinfo{year}{2021}\natexlab{}.
\newblock \showarticletitle{Hybrid multitask multi-information fusion deep learning for household short-term load forecasting}.
\newblock \bibinfo{journal}{\emph{IEEE Transactions on Smart Grid}} \bibinfo{volume}{12}, \bibinfo{number}{6} (\bibinfo{year}{2021}), \bibinfo{pages}{5362--5372}.
\newblock


\bibitem[Ke et~al\mbox{.}(2017)]%
        {ke2017lightgbm}
\bibfield{author}{\bibinfo{person}{Guolin Ke}, \bibinfo{person}{Qi Meng}, \bibinfo{person}{Thomas Finley}, \bibinfo{person}{Taifeng Wang}, \bibinfo{person}{Wei Chen}, \bibinfo{person}{Weidong Ma}, \bibinfo{person}{Qiwei Ye}, {and} \bibinfo{person}{Tie-Yan Liu}.} \bibinfo{year}{2017}\natexlab{}.
\newblock \showarticletitle{LightGBM: A Highly Efficient Gradient Boosting Decision Tree}. In \bibinfo{booktitle}{\emph{Advances in Neural Information Processing Systems}}, \bibfield{editor}{\bibinfo{person}{I.~Guyon}, \bibinfo{person}{U.~Von Luxburg}, \bibinfo{person}{S.~Bengio}, \bibinfo{person}{H.~Wallach}, \bibinfo{person}{R.~Fergus}, \bibinfo{person}{S.~Vishwanathan}, {and} \bibinfo{person}{R.~Garnett}} (Eds.), Vol.~\bibinfo{volume}{30}. \bibinfo{publisher}{Curran Associates, Inc.}
\newblock
\urldef\tempurl%
\url{https://proceedings.neurips.cc/paper/2017/file/6449f44a102fde848669bdd9eb6b76fa-Paper.pdf}
\showURL{%
\tempurl}


\bibitem[Kingma and Ba(2014)]%
        {kingma2014adam}
\bibfield{author}{\bibinfo{person}{Diederik~P Kingma} {and} \bibinfo{person}{Jimmy Ba}.} \bibinfo{year}{2014}\natexlab{}.
\newblock \showarticletitle{Adam: A method for stochastic optimization}.
\newblock \bibinfo{journal}{\emph{arXiv preprint arXiv:1412.6980}} (\bibinfo{year}{2014}).
\newblock


\bibitem[Kipf and Welling(2016)]%
        {kipf2016semi}
\bibfield{author}{\bibinfo{person}{Thomas~N Kipf} {and} \bibinfo{person}{Max Welling}.} \bibinfo{year}{2016}\natexlab{}.
\newblock \showarticletitle{Semi-supervised classification with graph convolutional networks}.
\newblock \bibinfo{journal}{\emph{arXiv preprint arXiv:1609.02907}} (\bibinfo{year}{2016}).
\newblock


\bibitem[Kuster et~al\mbox{.}(2017)]%
        {ELF:2017:survey}
\bibfield{author}{\bibinfo{person}{Corentin Kuster}, \bibinfo{person}{Yacine Rezgui}, {and} \bibinfo{person}{Monjur Mourshed}.} \bibinfo{year}{2017}\natexlab{}.
\newblock \showarticletitle{Electrical load forecasting models: A critical systematic review}.
\newblock \bibinfo{journal}{\emph{Sustainable Cities and Society}}  \bibinfo{volume}{35} (\bibinfo{year}{2017}), \bibinfo{pages}{257--270}.
\newblock
\showISSN{2210-6707}
\href{https://doi.org/10.1016/j.scs.2017.08.009}{doi:\nolinkurl{10.1016/j.scs.2017.08.009}}


\bibitem[Liu et~al\mbox{.}(2021)]%
        {liu2021conflict}
\bibfield{author}{\bibinfo{person}{Bo Liu}, \bibinfo{person}{Xingchao Liu}, \bibinfo{person}{Xiaojie Jin}, \bibinfo{person}{Peter Stone}, {and} \bibinfo{person}{Qiang Liu}.} \bibinfo{year}{2021}\natexlab{}.
\newblock \showarticletitle{Conflict-averse gradient descent for multi-task learning}.
\newblock \bibinfo{journal}{\emph{Advances in Neural Information Processing Systems}}  \bibinfo{volume}{34} (\bibinfo{year}{2021}), \bibinfo{pages}{18878--18890}.
\newblock


\bibitem[Liu et~al\mbox{.}(2023)]%
        {liu2023boosted}
\bibfield{author}{\bibinfo{person}{Haizhou Liu}, \bibinfo{person}{Xuan Zhang}, \bibinfo{person}{Hongbin Sun}, {and} \bibinfo{person}{Mohammad Shahidehpour}.} \bibinfo{year}{2023}\natexlab{}.
\newblock \showarticletitle{Boosted multi-task learning for inter-district collaborative load forecasting}.
\newblock \bibinfo{journal}{\emph{IEEE Transactions on Smart Grid}} \bibinfo{volume}{15}, \bibinfo{number}{1} (\bibinfo{year}{2023}), \bibinfo{pages}{973--986}.
\newblock


\bibitem[Lüpeng et~al\mbox{.}(2018)]%
        {electrician-competition}
\bibfield{author}{\bibinfo{person}{CHEN Lüpeng}, \bibinfo{person}{YIN Linfei}, {and} \bibinfo{person}{et~al. YU~Tao}.} \bibinfo{year}{2018}\natexlab{}.
\newblock \showarticletitle{Short-term Power Load Forecasting Based on Deep Forest Algorithm[J].}
\newblock \bibinfo{journal}{\emph{Electric Power Construction}} \bibinfo{volume}{39}, \bibinfo{number}{11} (\bibinfo{year}{2018}), \bibinfo{pages}{42--50}.
\newblock
\href{https://doi.org/10.3969/j.issn.1000-7229.2018.11.006}{doi:\nolinkurl{10.3969/j.issn.1000-7229.2018.11.006}}


\bibitem[L’Heureux et~al\mbox{.}(2022)]%
        {l2022transformer}
\bibfield{author}{\bibinfo{person}{Alexandra L’Heureux}, \bibinfo{person}{Katarina Grolinger}, {and} \bibinfo{person}{Miriam~AM Capretz}.} \bibinfo{year}{2022}\natexlab{}.
\newblock \showarticletitle{Transformer-based model for electrical load forecasting}.
\newblock \bibinfo{journal}{\emph{Energies}} \bibinfo{volume}{15}, \bibinfo{number}{14} (\bibinfo{year}{2022}), \bibinfo{pages}{4993}.
\newblock


\bibitem[Mansoor et~al\mbox{.}(2024)]%
        {mansoor2024graph}
\bibfield{author}{\bibinfo{person}{Haris Mansoor}, \bibinfo{person}{Muhammad~Shuzub Gull}, \bibinfo{person}{Huzaifa Rauf}, \bibinfo{person}{Muhammad Khalid}, \bibinfo{person}{Naveed Arshad}, {et~al\mbox{.}}} \bibinfo{year}{2024}\natexlab{}.
\newblock \showarticletitle{Graph Convolutional Networks based short-term load forecasting: Leveraging spatial information for improved accuracy}.
\newblock \bibinfo{journal}{\emph{Electric Power Systems Research}}  \bibinfo{volume}{230} (\bibinfo{year}{2024}), \bibinfo{pages}{110263}.
\newblock


\bibitem[Mounir et~al\mbox{.}(2023)]%
        {mounir2023short}
\bibfield{author}{\bibinfo{person}{Nada Mounir}, \bibinfo{person}{Hamid Ouadi}, {and} \bibinfo{person}{Ismael Jrhilifa}.} \bibinfo{year}{2023}\natexlab{}.
\newblock \showarticletitle{Short-term electric load forecasting using an EMD-BI-LSTM approach for smart grid energy management system}.
\newblock \bibinfo{journal}{\emph{Energy and Buildings}}  \bibinfo{volume}{288} (\bibinfo{year}{2023}), \bibinfo{pages}{113022}.
\newblock


\bibitem[Oreshkin et~al\mbox{.}(2021)]%
        {oreshkin2021n}
\bibfield{author}{\bibinfo{person}{Boris~N Oreshkin}, \bibinfo{person}{Grzegorz Dudek}, \bibinfo{person}{Pawe{\l} Pe{\l}ka}, {and} \bibinfo{person}{Ekaterina Turkina}.} \bibinfo{year}{2021}\natexlab{}.
\newblock \showarticletitle{N-BEATS neural network for mid-term electricity load forecasting}.
\newblock \bibinfo{journal}{\emph{Applied Energy}}  \bibinfo{volume}{293} (\bibinfo{year}{2021}), \bibinfo{pages}{116918}.
\newblock


\bibitem[Salam and Hibaoui(2018)]%
        {power-consumption-prediction}
\bibfield{author}{\bibinfo{person}{A. Salam} {and} \bibinfo{person}{A.~E. Hibaoui}.} \bibinfo{year}{2018}\natexlab{}.
\newblock \showarticletitle{Comparison of Machine Learning Algorithms for the Power Consumption Prediction: Case Study of Tetouan city}.
\newblock \bibinfo{journal}{\emph{2018 6th International Renewable and Sustainable Energy Conference (IRSEC)}} (\bibinfo{year}{2018}).
\newblock
\href{https://doi.org/10.1109/irsec.2018.8703007}{doi:\nolinkurl{10.1109/irsec.2018.8703007}}


\bibitem[Shumway et~al\mbox{.}(2017)]%
        {shumway2017arima}
\bibfield{author}{\bibinfo{person}{Robert~H Shumway}, \bibinfo{person}{David~S Stoffer}, \bibinfo{person}{Robert~H Shumway}, {and} \bibinfo{person}{David~S Stoffer}.} \bibinfo{year}{2017}\natexlab{}.
\newblock \showarticletitle{ARIMA models}.
\newblock \bibinfo{journal}{\emph{Time series analysis and its applications: with R examples}} (\bibinfo{year}{2017}), \bibinfo{pages}{75--163}.
\newblock


\bibitem[Taylor and Letham(2018)]%
        {taylor2018forecasting}
\bibfield{author}{\bibinfo{person}{Sean~J Taylor} {and} \bibinfo{person}{Benjamin Letham}.} \bibinfo{year}{2018}\natexlab{}.
\newblock \showarticletitle{Forecasting at scale}.
\newblock \bibinfo{journal}{\emph{The American Statistician}} \bibinfo{volume}{72}, \bibinfo{number}{1} (\bibinfo{year}{2018}), \bibinfo{pages}{37--45}.
\newblock


\bibitem[Wen et~al\mbox{.}(2019)]%
        {RobustSTL_wen2018robuststl}
\bibfield{author}{\bibinfo{person}{Qingsong Wen}, \bibinfo{person}{Jingkun Gao}, \bibinfo{person}{Xiaomin Song}, \bibinfo{person}{Liang Sun}, \bibinfo{person}{Huan Xu}, {and} \bibinfo{person}{Shenghuo Zhu}.} \bibinfo{year}{2019}\natexlab{}.
\newblock \showarticletitle{RobustSTL: A robust seasonal-trend decomposition algorithm for long time series}. In \bibinfo{booktitle}{\emph{AAAI}}. \bibinfo{pages}{1501--1509}.
\newblock


\bibitem[Wen et~al\mbox{.}(2021)]%
        {wen2021robustperiod}
\bibfield{author}{\bibinfo{person}{Qingsong Wen}, \bibinfo{person}{Kai He}, \bibinfo{person}{Liang Sun}, \bibinfo{person}{Yingying Zhang}, \bibinfo{person}{Min Ke}, {and} \bibinfo{person}{Huan Xu}.} \bibinfo{year}{2021}\natexlab{}.
\newblock \showarticletitle{RobustPeriod: Robust Time-Frequency Mining for Multiple Periodicity Detection}. In \bibinfo{booktitle}{\emph{Proceedings of the 2021 International Conference on Management of Data(SIGMOD)}}. \bibinfo{pages}{2328--2337}.
\newblock


\bibitem[Wen et~al\mbox{.}(2020)]%
        {FastRobustSTL_wen2020}
\bibfield{author}{\bibinfo{person}{Qingsong Wen}, \bibinfo{person}{Zhe Zhang}, \bibinfo{person}{Yan Li}, {and} \bibinfo{person}{Liang Sun}.} \bibinfo{year}{2020}\natexlab{}.
\newblock \showarticletitle{{Fast RobustSTL}: Efficient and Robust Seasonal-Trend Decomposition for Time Series with Complex Patterns}. In \bibinfo{booktitle}{\emph{Proceedings of the 26th ACM SIGKDD International Conference on Knowledge Discovery \& Data Mining (KDD)}}. \bibinfo{pages}{2203--2213}.
\newblock


\bibitem[Wu et~al\mbox{.}(2022)]%
        {Wu:MRELF:2022}
\bibfield{author}{\bibinfo{person}{Kailang Wu}, \bibinfo{person}{Jie Gu}, \bibinfo{person}{Lu Meng}, \bibinfo{person}{Honglin Wen}, {and} \bibinfo{person}{Jinghuan Ma}.} \bibinfo{year}{2022}\natexlab{}.
\newblock \showarticletitle{An explainable framework for load forecasting of a regional integrated energy system based on coupled features and multi-task learning}.
\newblock \bibinfo{journal}{\emph{Protection and Control of Modern Power Systems}} \bibinfo{volume}{7}, \bibinfo{number}{2} (\bibinfo{year}{2022}), \bibinfo{pages}{1--14}.
\newblock
\href{https://doi.org/10.1186/s41601-022-00245-y}{doi:\nolinkurl{10.1186/s41601-022-00245-y}}


\bibitem[Yang et~al\mbox{.}(2024a)]%
        {yang2024multi}
\bibfield{author}{\bibinfo{person}{Yuqi Yang}, \bibinfo{person}{Peng-Tao Jiang}, \bibinfo{person}{Qibin Hou}, \bibinfo{person}{Hao Zhang}, \bibinfo{person}{Jinwei Chen}, {and} \bibinfo{person}{Bo Li}.} \bibinfo{year}{2024}\natexlab{a}.
\newblock \showarticletitle{Multi-Task Dense Prediction via Mixture of Low-Rank Experts}. In \bibinfo{booktitle}{\emph{Proceedings of the IEEE/CVF Conference on Computer Vision and Pattern Recognition}}. \bibinfo{pages}{27927--27937}.
\newblock


\bibitem[Yang et~al\mbox{.}(2024b)]%
        {MLoRE}
\bibfield{author}{\bibinfo{person}{Yuqi Yang}, \bibinfo{person}{Peng-Tao Jiang}, \bibinfo{person}{Qibin Hou}, \bibinfo{person}{Hao Zhang}, \bibinfo{person}{Jinwei Chen}, {and} \bibinfo{person}{Bo Li}.} \bibinfo{year}{2024}\natexlab{b}.
\newblock \showarticletitle{Multi-Task Dense Prediction via Mixture of Low-Rank Experts}. In \bibinfo{booktitle}{\emph{Proceedings of the IEEE/CVF Conference on Computer Vision and Pattern Recognition (CVPR)}}. \bibinfo{pages}{27927--27937}.
\newblock


\bibitem[Ye and Xu(2023)]%
        {ye2023taskexpert}
\bibfield{author}{\bibinfo{person}{Hanrong Ye} {and} \bibinfo{person}{Dan Xu}.} \bibinfo{year}{2023}\natexlab{}.
\newblock \showarticletitle{Taskexpert: Dynamically assembling multi-task representations with memorial mixture-of-experts}. In \bibinfo{booktitle}{\emph{Proceedings of the IEEE/CVF International Conference on Computer Vision}}. \bibinfo{pages}{21828--21837}.
\newblock


\bibitem[Zhaoyang et~al\mbox{.}(2023)]%
        {eForecaster2023}
\bibfield{author}{\bibinfo{person}{Zhu Zhaoyang}, \bibinfo{person}{Chen Weiqi}, \bibinfo{person}{Xia Rui}, \bibinfo{person}{Zhou Tian}, \bibinfo{person}{Niu Peisong}, \bibinfo{person}{Peng Bingqing}, \bibinfo{person}{Wang Wenwei}, \bibinfo{person}{Liu Hengbo}, \bibinfo{person}{Ma Ziqing}, \bibinfo{person}{Wen Qingsong}, {and} \bibinfo{person}{Sun Liang}.} \bibinfo{year}{2023}\natexlab{}.
\newblock \showarticletitle{eForecaster: Unifying Electricity Forecasting with Robust, Flexible, and Explainable Machine Learning Algorithms}. In \bibinfo{booktitle}{\emph{Thirty-Seventh AAAI Conference on Artificial Intelligence}}. AAAI.
\newblock


\bibitem[Zhou et~al\mbox{.}(2021)]%
        {zhou2021informer}
\bibfield{author}{\bibinfo{person}{Haoyi Zhou}, \bibinfo{person}{Shanghang Zhang}, \bibinfo{person}{Jieqi Peng}, \bibinfo{person}{Shuai Zhang}, \bibinfo{person}{Jianxin Li}, \bibinfo{person}{Hui Xiong}, {and} \bibinfo{person}{Wancai Zhang}.} \bibinfo{year}{2021}\natexlab{}.
\newblock \showarticletitle{Informer: Beyond efficient transformer for long sequence time-series forecasting}. In \bibinfo{booktitle}{\emph{Proceedings of the AAAI conference on artificial intelligence}}, Vol.~\bibinfo{volume}{35}. \bibinfo{pages}{11106--11115}.
\newblock


\bibitem[Zhou et~al\mbox{.}(2022)]%
        {zhou2022fedformer}
\bibfield{author}{\bibinfo{person}{Tian Zhou}, \bibinfo{person}{Ziqing Ma}, \bibinfo{person}{Qingsong Wen}, \bibinfo{person}{Xue Wang}, \bibinfo{person}{Liang Sun}, {and} \bibinfo{person}{Rong Jin}.} \bibinfo{year}{2022}\natexlab{}.
\newblock \showarticletitle{Fedformer: Frequency enhanced decomposed transformer for long-term series forecasting}. In \bibinfo{booktitle}{\emph{International Conference on Machine Learning}}. PMLR, \bibinfo{pages}{27268--27286}.
\newblock


\bibitem[Zhu et~al\mbox{.}(2023)]%
        {zhu2023eforecaster}
\bibfield{author}{\bibinfo{person}{Zhaoyang Zhu}, \bibinfo{person}{Weiqi Chen}, \bibinfo{person}{Rui Xia}, \bibinfo{person}{Tian Zhou}, \bibinfo{person}{Peisong Niu}, \bibinfo{person}{Bingqing Peng}, \bibinfo{person}{Wenwei Wang}, \bibinfo{person}{Hengbo Liu}, \bibinfo{person}{Ziqing Ma}, \bibinfo{person}{Qingsong Wen}, {and} \bibinfo{person}{Liang Sun}.} \bibinfo{year}{2023}\natexlab{}.
\newblock \showarticletitle{{eForecaster}: unifying electricity forecasting with robust, flexible, and explainable machine learning algorithms}. In \bibinfo{booktitle}{\emph{Proceedings of the AAAI Conference on Artificial Intelligence}}, Vol.~\bibinfo{volume}{37}. \bibinfo{pages}{15630--15638}.
\newblock


\end{thebibliography}


\newpage
\appendix
\onecolumn

\section{Appendix}

\subsection{Baseline Specifications and Implementation Details}
\label{appendix:backbone_baseline_details}

\subsubsection{Baseline Specifications}
\begin{itemize}
    \item \textbf{STL} trains models independently for each region, utilizing an identical architecture for each. Specifically, each model is TSmixer~\citep{chen2023tsmixer} with six blocks and two-layer MLP prediction head.
    \item \textbf{MTL} employs a unified TSmixer backbone consisting of six blocks, shared across all regions. In line with conventional multitask learning paradigms, this architecture includes region-specific prediction heads achieved through independently parameterized two-layer MLPs for each region.
    \item \textbf{MLoRE}~\citep{MLoRE} introduces a decoder-centric architecture for multi-task dense prediction, originally developed for visual domain applications. This SOTA MoE approach enhances the efficiency of expert networks through low-rank convolutional decompositions. In our adaptation, we replace the original backbone with a TSmixer consisting of six blocks.
    \item \textbf{GCN-LSTM}~\citep{mansoor2024graph} is a SOTA MRELF method that employs GCN~\citep{kipf2016semi} for spatial and LSTM~\citep{hochreiter1997long} for temporal feature extraction. In our adaptation, we replace LSTM with a six-block TSmixer. The model first extracts temporal features via Tsmixer, then propagates these representations through the GCN to aggregate spatial information from geographically adjacent regions connected by undirected edges in the graph topology.
\end{itemize}

\subsubsection{Implementation Details}
\label{appendix:implementation_details}

For all benchmarks, we set the latent dimension $d_r$ to 16. We utilize an early stopping strategy during model training. For the EPC, CEESC, and Bus-load datasets, we set the maximum number of epochs to 200 and a patience of 50 to ensure efficient training. In contrast, for the City-load dataset, given the large data volume and the number of cities involved, we set the maximum number of epochs to 50 and a patience of 10.

\subsection{Studies of Hyperparameter Sensitivity}
\label{appendix:hyper_sen}
\begin{figure}[h]
    \centering
    \includegraphics[width=0.8\linewidth]{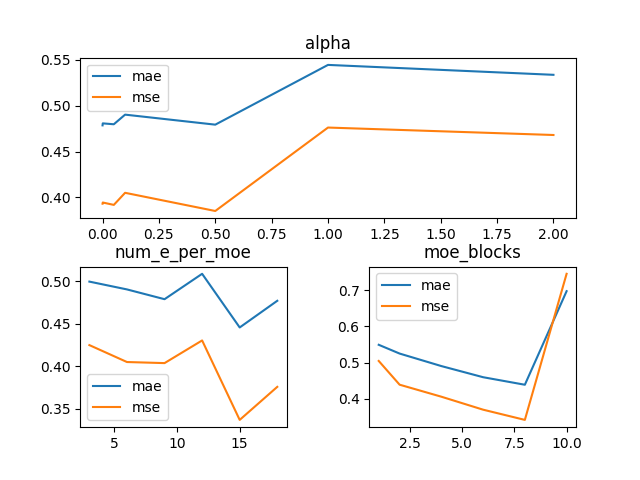}
    \caption{The effect of hyperparameters.}
    \label{fig:hyperparam_sen}
\end{figure}

As shown in Figure \ref{fig:hyperparam_sen}, we run sensitivity experiments for three major hyperparameters: (1) \textbf{alpha.} The coefficient of contrastive loss; (2) \textbf{num\_e\_per\_moe.} The number of experts in each CTSpecializer layer; (3) \textbf{moe\_blocks.} The number of CTSpecializer layers. The effectiveness of the contrastive loss is demonstrated, as the error gradualy decreases as we increase the coefficient. Notably, model performance can also be enhanced by merely increasing the number of CTSpecailizer layers and the number of experts.

\begin{table}[h]
\caption{Standard deviations of MSE in Table \ref{table:comparison_performance}.
}
\label{table:std}
\vspace{-2mm}
\normalsize
\centering
\renewcommand\arraystretch{0.8}
\begin{tabular}{c| c| c| c | c}
\toprule
Method & EPC & CEESC & City-load & Bus-load \\
\midrule
STL & 0.1675 & 0.0538 & 0.0738  & 0.2856   \\
MTL &  0.0274 &  0.0308 &  0.0151  &  0.0261   \\
 MLoRE & {0.0127} & 0.0246 & 0.0067  & {0.0346}   \\
 GCN & 0.0826 & 0.0504 & 0.0021 & 0.0170   \\
 \midrule
 TriForecaster(ours) & 0.0086 &	0.0041	&  0.0011 &	0.0180   \\

\bottomrule
\end{tabular}

\end{table}

\end{document}